\documentclass[12pt]{article}
\usepackage{epsf}
\def\paragraph#1{{\bf #1}}
\newenvironment{keywords}{\centerline{\bf
Keywords}\vspace{0.5ex}\begin{quote}\small}{\par\end{quote}\vskip
1ex}

\def\gtapprox{\buildrel{\lower.7ex\hbox{$>$}}\over
                       {\lower.7ex\hbox{$\sim$}}}

\def\ignore#1{}

\def\odt{{\textstyle{1\over 2}}}

\def\hbar{h\!\!\!\!^{-}\,}

\def\eps{\varepsilon}
\def\beq{\begin{equation}}
\def\eeq{\end{equation}}
\def\beqn{\begin{displaymath}}
\def\eeqn{\end{displaymath}}
\def\bqa{\begin{equation}\begin{array}{c}}
\def\eqa{\end{array}\end{equation}}
\def\bqan{\begin{displaymath}\begin{array}{c}}
\def\eqan{\end{array}\end{displaymath}}
\def\vec#1{{\bf #1}}
\def\approxleq{\mbox{\raisebox{-0.8ex}{$\stackrel{\displaystyle<}\sim$}}} 

\begin{document}

\begin{titlepage}

\begin{center}
    {\small Technical Report IDSIA-01-01, 17 January 2001 \\
            {ftp://ftp.idsia.ch/pub/techrep/IDSIA-01-01.ps.gz}
    }\\[2cm]
  {\Huge\sc\sc\hrule height1pt \vskip 2mm
          Fitness Uniform Selection to \\ Preserve Genetic Diversity
   \vskip 5mm \hrule height1pt} \vspace{1.5cm}
  {\bf Marcus Hutter}                    \\[1cm]
  {\rm IDSIA, Galleria 2, CH-6928 Manno-Lugano, Switzerland}  \\
  {\rm\footnotesize marcus@idsia.ch \qquad
      http://www.idsia.ch/$^{_{_\sim}}\!$marcus} \\[1cm]
\end{center}

\begin{keywords}
Evolutionary algorithms, fitness uniform selection strategy,
preserve diversity, local optima, evolution,
correlated recombination, crossover.
\end{keywords}

\begin{abstract}
In evolutionary algorithms, the fitness of a population increases
with time by mutating and recombining individuals and by a biased
selection of more fit individuals. The right selection pressure is
critical in ensuring sufficient optimization progress on the one
hand and in preserving genetic diversity to be able to escape from
local optima on the other. We propose a new selection scheme,
which is uniform in the fitness values. It generates selection
pressure towards sparsely populated fitness regions, not
necessarily towards higher fitness, as is the case for all other
selection schemes. We show that the new selection scheme can be
much more effective than standard selection schemes.
\end{abstract}

\end{titlepage}

\section{Introduction}\label{secInt}

\paragraph{Evolutionary algorithms (EA):}
Evolutionary algorithms are capable of solving complicated
optimization tasks in which an objective function
$f\!:\!I\!\to\!I\!\!R$ shall be maximized. $i\!\in\!I$ is an
individual from the set $I$ of feasible solutions. Infeasible
solutions due to constraints my also be considered by reducing
$f$ for each violated constraint. A population $P\!\subseteq\!I$
of individuals is maintained and updated as follows: one or more
individuals are selected according to some selection strategy.
In generation based EAs, the selected individuals are recombined
(e.g.\ crossover) and mutated, and constitute the new population.
We prefer the more incremental, steady-state population update,
which selects (and possibly delets) only one or two individuals from
the current population and adds the newly recombined and mutated
individuals to it.
We are interested in finding a single individual of maximal objective value
$f$ for difficult multimodal and deceptive problems.

\paragraph{Standard selection schemes (STD):}
The standard selection schemes (abbreviated by STD in the
following), proportionate, truncation, ranking and tournament
selection all favor individuals of higher fitness
\cite{Goldberg:89,Goldberg:91,Blickle:95a,Blickle:97}. This is
also true for less common schemes, like Boltzmann selection
\cite{Maza:93}. The
fitness function is identified with the objective
function (possibly after a monotone transformation).
In linear proportionate selection the probability
of selecting an individual depends linearly on its fitness
\cite{Holland:75}. In truncation selection the $\alpha\%$ fittest
individuals are selected, usually with multiplicity
${1\over\alpha\%}$ in order to keep the population size fixed
\cite{Muehlenbein:94}. (Linear) ranking selection orders the
individuals according to their fitness. The selection probability
is, then, a (linear) function of the rank \cite{Whitley:89}.
Tournament selection \cite{Baker:85}, which selects the best $l$
out of $k$ individuals has primarily developed for steady-state
EAs, but can be adapted to generation based EAs. All these
selection schemes have the property (and goal!) to increase the
average fitness of a population, i.e.\ to evolve the population
towards higher fitness. For a population with a Gaussian fitness
distribution, the probability of selecting an individual and the
effect of selection is shown in Figure \ref{figsel}.

\paragraph{The problem of the right selection pressure:}
The standard selection schemes STD, together with mutation and
recombination, evolve the population towards higher fitness. If
the selection pressure is too high, the EA gets stuck in a local
optimum, since the genetic diversity rapidly decreases. The
suboptimal genetic material which might help in finding the global
optimum is deleted too rapidly (premature convergence). On the other hand, the selection
pressure cannot be chosen arbitrarily low if we want EA to be
effective. In difficult optimization problems, suitable population
sizes, mutation and recombination rates, and selection parameters,
which influence the selection intensity, are usually not known
beforehand. Often, constant values are not sufficient at all.
There are various suggestions to dynamically determine and adapt
the parameters
\cite{Eshelman:91,Baeck:91,Herdy:92,Schlierkamp:94}.
Other approaches to preserve genetic diversity are fitness sharing
\cite{Goldberg:87}, crowding \cite{DeJong:75} and local mating
\cite{Collins:91}. They depend on the proper design of a
neighborhood function based on the specific problem structure
and/or coding.

\paragraph{The main idea:}
Here, we propose a new selection scheme, based on the insight that
we are not primarily interested in a population converging to
maximal fitness, but only in a single individual of maximal fitness.
The scheme automatically
creates a suitable selection pressure and preserves genetic
diversity better than STD. The proposed fitness uniform selection
scheme FUSS (see also Figure \ref{figsel}) is defined as follows:
{\em if the lowest/highest fitness values in the current
population $P$ are $f_{min/max}$, we select a fitness value $f$
uniformly in the interval $[f_{min},f_{max}]$. Then, the
individual $i\!\in\!P$ with fitness nearest to $f$ is selected and
a copy is added to $P$, possibly after mutation and
recombination.} We will see that FUSS maintains genetic diversity
much better than STD, since a distribution over the fitness values
is used, unlike STD, which all use a distribution over
individuals. Premature convergence is avoided in FUSS by
abandoning convergence at all. Nevertheless there is a selection
pressure in FUSS towards higher fitness.
The probability of selecting a specific individual is proportional
to the distance to its nearest fitness neighbor. In a
population with a high density of unfit and low density of fit
individuals, the fitter ones are effectively favored.

\paragraph{Contents:}
In {\em Section \ref{secFuss}} we discuss the problems of local
optima and exponential takeover \cite{Goldberg:91} in STD.
Motivated by the need to preserve genetic diversity, we define the
fitness uniform selection scheme FUSS. We discuss under which
circumstances FUSS leads to an (approximate) fitness uniform
population.

Further properties of FUSS are discussed in {\em Section
\ref{secProp}}, especially, how FUSS creates selection pressure
towards higher fitness and how it preserves diversity better than
STD. Further topics are the equilibrium distribution and the
transformation properties of FUSS under linear and non-linear
transformations.

In {\em Section \ref{secEx}} we demonstrate by way of a simple
optimization example that an EA with FUSS can optimize much faster
than with STD. We show that crossover can be effective
in FUSS, even when ineffective in STD. Furthermore, FUSS and STD are
compared to random search with and without crossover.

There is a possible slowdown when including recombination, as
discussed in {\em Section \ref{secCross}}, which can be
avoided by using a scale independent pair selection. It is a
``best'' compromise between unrestricted recombination and
recombination of individuals with similar fitness only.

To simplify the discussion we have concentrated on the case of
discrete, equi-spaced fitness values. In many practical problems,
the fitness function is continuously valued. FUSS and some of the
discussion of the previous sections is generalized to the
continuous case in {\em Section \ref{secCont}}.

A summary, conclusions and further discussions can be found in
{\em Section \ref{secConc}}.

The focus of this work is on a theoretical analysis of FUSS.
Implementation details and numerical results for various
test-functions and for real-world problems will be presented
elsewhere.

\section{Fitness Uniform Selection Strategy (FUSS)}\label{secFuss}

\paragraph{The problem of local optima:}
Proportionate, truncation, ranking and tournament are the standard
(STD) selection algorithms used in evolutionary optimization. They
have the following property: if a local optimum $i^{lopt}$ has
been found, the number of individuals with fitness
$f^{lopt}=f(i^{lopt})$ increases exponentially. Assume a low
mutation and recombination rate, or, for instance, truncation
selection {\em after} mutation and recombination. Further, assume
that it is very difficult to find an individual more fit than
$i^{lopt}$. The population will then degenerate and will consist
mostly of $i^{lopt}$ after a few rounds. This decreased diversity
makes it even more unlikely that $f^{lopt}$ gets improved. The
suboptimal genetic material which might help in finding the global
optimum has been deleted too rapidly. On the other hand, too high
mutation and recombination rates convert the EA into an
inefficient random search. In the following we suggest a new
a new selection scheme, which automatically
generates a suitably adapting selection pressure.

\paragraph{The fitness uniform selection scheme (FUSS):}
For simplicity we start with a fitness function $f\!:\!I\!\to\!F$
with discrete equi-spaced values
$F\!=\!\{f_{min},f_{min}\!+\!\eps,f_{min}\!+\!2\eps,...,
f_{max}\!-\!\eps,f_{max}\}$. The continuous valued case
$F\!=\![f_{min},f_{max}]$ is considered later.
The fitness uniform selection scheme (FUSS) is defined as follows:
randomly select a fitness value $f$ uniformly from the fitness
values $F$. Randomly (uniformly) select an individual $i$ from
population $P$ with fitness $f$. Add another copy of $i$ to $P$.

\begin{figure}[tb]\epsfxsize=9cm
\centerline{\epsfbox{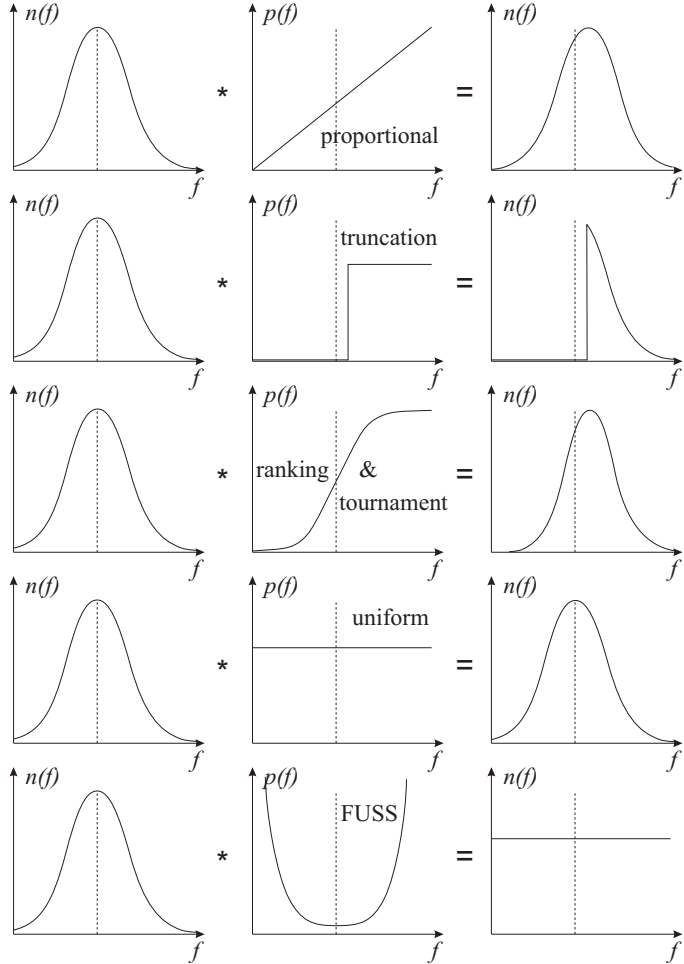}} 
\caption{\label{figsel}\it Effects of proportionate, truncation,
ranking \& tournament, uniform, and fitness uniform (FUSS)
selection on the fitness distribution in a generation based EA.
The left/right diagrams depict fitness distributions before/after
applying the selection schemes depicted in the middle diagrams.}
\end{figure}

Note the two stage uniform selection process which is very
different from a one step uniform selection of an individual of
$P$ (see Figure \ref{figsel}).
%
In STD, inertia increases with population size. A large mass of
unfit individuals reduces the probability of selecting fit
individuals. This is not the case for FUSS. Hence, without loss of
performance, we can define a {\em pure model}, in which no
individual is ever deleted; the population size increases with
time. No genetic material is ever discarded and no fine-tuning in
population size is necessary. What may prevent the pure model from
being applied to practical problems are not computation time
issues, but memory problems. If space gets a problem we delete
individuals from the most occupied fitness levels. Most of the
following statements remain valid with this modification.

\paragraph{Asymptotically fitness uniform population:}
The expected number of
individuals per fitness level $f$ after $t$ selections is
$n_t(f)=n_0(f)+t/|F|$, where $n_0(f)$ is the initial distribution.
Hence, asymptotically the fitness levels get uniformly occupied
by a population fraction
$$
  {n_t(f)\over |P_t(f)|} \;=\;
  {n_0(f)+t/|F|\over |P_0(f)|+t} \;\to\; {1\over |F|}
  \quad\mbox{for}\quad t\to\infty,
$$
where $P_t(f)$ is the set of individuals at time $t$ with fitness
$f$.

\paragraph{Fitness gaps and continuous fitness:}
We made two unrealistic assumptions. First, we assumed that each
fitness level is initially occupied. If the smallest/largest
fitness values in $P_t$ are $f_{min/max}^t$ we extend the
definition of FUSS by selecting a fitness value $f$ uniformly in
the interval $[f_{min}^t-\odt\eps,f_{max}^t+\odt\eps]$ and an
individual $i\!\in\!P_t$ with fitness nearest to $f$. This also
covers the case when there are missing intermediate fitness
values, and also works for continuous valued fitness functions
($\eps\to 0$).

\paragraph{Mutation and recombination:}
The second assumption was that there is no mutation and
recombination. In the presence of small mutation and/or
recombination rates eventually each fitness level will become
occupied and the occupation fraction is still asymptotically
approximately uniform. For larger rates the distribution will be no
longer uniform, but the important point is that the occupation
fraction of {\em no} fitness level decreases to zero for
$t\!\to\!\infty$, unlike for STD. Furthermore, FUSS selects by
construction uniformly in the fitness levels, even if the levels
are not uniformly occupied. We will see that this is the more
important property.

\section{Properties of Fuss}\label{secProp}
\paragraph{FUSS effectively favors fit individuals:}
FUSS preserves diversity better than STD, but the latter have a
(higher) selection pressure towards higher fitness, which is
necessary for optimization. At first glance it seems that there is
no such pressure at all in FUSS, but this is deceiving. As FUSS selects
uniformly in the fitness levels, individuals of low populated
fitness levels are effectively favored. The probability of
selecting a specific individual with fitness $f$ is inverse
proportional to $n_t(f)$ (see Figure \ref{figsel}). In a typical
(FUSS) population there are many unfit and only a few fit
individuals. Hence, fit individuals are effectively favored until
the population becomes fitness uniform. Occasionally, a new higher
fitness level is discovered and occupied by a new individual,
which then, again, is favored.

\paragraph{No takeover in FUSS:}
With FUSS, takeover of the highest fitness level never happens.
The concept of takeover time \cite{Goldberg:91} is meaningless for
FUSS. The fraction of fittest individuals in a population is
always small. This implies that the average population fitness is
always much lower than the best fitness. Actually, a large number
of fit individuals is usually not the true optimization goal. A
single fittest individual usually suffices to having solved the
optimization task.

\paragraph{FUSS may also favor unfit individuals:}
Note, if it is also difficult to find individuals of low fitness,
i.e.\ if there are only few individuals of low fitness, FUSS will
also favor these individuals. Half of the time is ``wasted'' in
searching on the wrong end of the fitness scale. This possible
slowdown by a factor of 2 is usually acceptable. In Section
\ref{secEx} we will see that in certain circumstances this
behaviour can actually speedup the search. In general, fitness
levels which are difficult to reach, are favored.

\paragraph{Distribution within a fitness level:}
Within a fitness level there is no selection pressure which could
further exponentially decrease the population in certain regions
of the individual space. This (exponential) reduction is the major
enemy of diversity, which is suppressed by FUSS. Within a fitness
level, the individuals freely drift around (by mutation).
Furthermore, there is a steady stream of individuals into and out
of a level by (d)evolution from (higher)lower levels.
Consequently, FUSS develops an equilibrium distribution which is
nowhere zero. We expect FUSS to somewhat lower (but not to solve)
the problems associated with genetic drift. The above does also
not mean that the distribution within a level is uniform. For
instance, if there are two (local) maxima of same height, a very
broad one and a very narrow one, the broad one may be populated
much more than the narrow one, since it is much easier to
``find''.

\paragraph{Steady creation of individuals from every fitness level:}
In STD, a wrong step (mutation) at some point in evolution might
cause further evolution in the wrong direction. Once a local
optimum has been found and all unfit individuals were eliminated
it is very difficult to undo the wrong step. In FUSS, all fitness
levels remain occupied from which new mutants are steadily
created, occasionally one leading to further evolution in a more
promising direction.

\paragraph{Transformation properties of FUSS:}
FUSS (with continuous fitness) is independent of a scaling and a
shift of the fitness function, i.e.\ FUSS($\tilde f$) with $\tilde
f(i):=a\!\cdot\!f(i)+b$ is identical to FUSS($f$). This is true
even for $a\!<\!0$, since FUSS searches for maxima {\em and}
minima, as we have seen. It is not independent of a non-linear
(monotone) transformation unlike tournament, ranking and
truncation selection. The non-linear transformation properties are
more like the ones of proportionate selection.

\section{A Simple Example}\label{secEx}

In the following we compare the performance of fitness uniform
selection (FUSS), random search (RAND) and standard selection
(STD) with and without recombination on a simple test example. We
regard it as a prototype for deceptive multimodal functions. The
example should demonstrate why FUSS can be superior to RAND and
STD. Numerical results are briefly discussed at the end of the
section.

\paragraph{Simple 2D example:}
Consider individuals $(x,y)\!\in\!I\!:=\![0,1]\!\times\![0,1]$,
which are tupels of real numbers, each coordinate in the interval $[0,1]$.
The example models individuals possessing up to 2 ``features''.
Individual $i$ possesses feature $I_1$ if
$i\!\in\!I_1:=[a,a\!+\!\Delta]\!\times\![0,1]$, and feature $I_2$
if $i\!\in\! I_2:=[0,1]\!\times\![b,b\!+\!\Delta]$.
The fitness function $f:I\!\to\!\{1,2,3\}$ is defined as
$$
  f(x,y) = \left\{
  \begin{array}{l}
    1 \quad\mbox{if}\quad (x,y)\in I_1\backslash I_2, \\
    2 \quad\mbox{if}\quad (x,y)\in I_2\backslash I_1, \\
    3 \quad\mbox{if}\quad (x,y)\not\in I_1\cup I_2, \\
    4 \quad\mbox{if}\quad (x,y)\in I_1\cap I_2. \\
  \end{array}\right.
\parbox{4cm}{\hfill \unitlength=0.7mm
\unitlength=0.7mm
\begin{picture}(45,45)
\footnotesize
\put(5,5){\vector(0,1){40}}
\put(5,5){\vector(1,0){40}}
\put(20,5){\line(0,1){35}}
\put(25,5){\line(0,1){35}}
\put(40,5){\line(0,1){35}}
\put(5,15){\line(1,0){35}}
\put(5,20){\line(1,0){35}}
\put(5,40){\line(1,0){35}}
\put(22.5,17.5){\makebox(0,0)[cc]{4}}
\put(12.5,30){\makebox(0,0)[cc]{3}}
\put(22.5,30){\makebox(0,0)[cc]{1}}
\put(32.5,30){\makebox(0,0)[cc]{3}}
\put(32.5,10){\makebox(0,0)[cc]{3}}
\put(12.5,10){\makebox(0,0)[cc]{3}}
\put(12.5,17.5){\makebox(0,0)[cc]{2}}
\put(32.5,17.5){\makebox(0,0)[cc]{2}}
\put(22.5,10){\makebox(0,0)[cc]{1}}
\put(44,2.5){\makebox(0,0)[cc]{$x$}}
\put(22.5,2.5){\makebox(0,0)[cc]{$\Delta$}}
\put(20,3.5){\makebox(0,0)[cc]{$a$}}
\put(2.5,17.5){\makebox(0,0)[cc]{$\Delta$}}
\put(4,14.5){\makebox(0,0)[cc]{$b$}}
\put(40,3){\makebox(0,0)[cc]{1}}
\put(3.5,40){\makebox(0,0)[cc]{1}}
\put(2.5,44){\makebox(0,0)[cc]{$y$}}
\put(22.5,42.5){\makebox(0,0)[cc]{$f(x,y)$}}
\end{picture} }
$$
We assume $\Delta\!\ll\!1$.
Individuals
with none of the two features
($i\!\in\!I\backslash(I_1\!\cup\!I_2)$) have fitness $f\!=\!3$.
These ``local $f\!=\!3$ optima'' occupy most of the individual
space $I$, namely a fraction $(1\!-\!\Delta)^2$. It is
disadvantageous for an individual to possess only one of the two
features ($i\!\in\!(I_1\backslash\!I_2)\cup
(I_2\backslash\!I_1)$), since $f\!=\!1/f\!=\!2$ in this case. In
combination ($i\!\in\!I_1\!\cap\!I_2)$), the two features lead to
the highest fitness, but the global maximum $f\!=\!4$ occupies the
smallest fraction $\Delta^2$ of the individual space $I$. With a
fraction $\Delta(1\!-\!\Delta)$, the $f\!=\!1/f\!=\!2$ minima are
in between.

\paragraph{Random search:}
Individuals are created uniformly in the unit square. The ``local
optimum'' $f\!=\!3$ is easy to ``find'', since it occupies nearly
the whole space. The global optimum $f\!=\!4$ is difficult to
find, since it occupies only $\Delta^2\!\ll\!1$ of the space. The
expected time, i.e.\ the expected number of individuals created
and tested until one with $f\!=\!4$ is found, is
$T_{RAND}={1\over\Delta^2}$. Here and in the following, the
``time'' $T$ is defined as the number of created individuals
until the {\it first} optimal individual (with
$f\!=\!4$) is found. $T$ is neither a takeover time nor the number
of generations (we consider steady-state EAs).

\paragraph{Random search with crossover:}
Let us occasionally perform a recombination of individuals in the
current population. We combine the $x$-coordinate of one uniformly
selected individual $i_1$ with the $y$ coordinate of another
individual $i_2$. This crossover operation maintains a uniform
distribution of individuals in $[0,1]^2$. It leads to the global
optimum if $i_1\!\in\!I_1$ and $i_2\!\in\!I_2$. The probability of
selecting an individual in $I_i$ is
$\Delta(1-\Delta)\approx\Delta$ (we assumed that the global
optimum has not yet been found). Hence, the probability that $I_1$
crosses with $I_2$ is $\Delta^2$. The time to find the global
optimum by random search including crossover is still
$\sim{1\over\Delta^2}$.

\paragraph{Mutation:}
The result remains valid (to leading order in ${1\over\Delta}$)
if, instead of a random search, we uniformly select an individual
and mutate it according to some probabilistic, sufficiently mixing
rule, which preserves uniformity in $[0,1]$. One popular such
mutation operator is to use a sufficiently long binary
representation of each coordinate, like in genetic algorithms, and
flip a single bit. In the following, we assume a mutation operator
which mutates with probability $\odt/\odt$ the first/second
coordinate, preserves uniformity, is sufficiently mixing, and
leaves the other coordinate unchanged, like the single-bit-flip
operator.

\paragraph{Standard selection with crossover:}
The $f\!=\!1$ and $f\!=\!2$ individuals contain useful building
blocks, which could speedup the search by a suitable selection and
crossover scheme. Unfortunately, the standard selection schemes
favor individuals of higher fitness and will diminish the
$f\!=\!1/f\!=\!2$ population fraction. The probability of
selecting $f\!=\!1/f\!=\!2$ individuals is even smaller than in
random search. Hence $T_{STD}\sim{1\over\Delta^2}$. Standard
selection does not improve performance, even not in combination
with crossover, although crossover is well suited to produce the
needed recombination.

\paragraph{FUSS:}
At the beginning, only the $f\!=\!3$ level is occupied and
individuals are uniformly selected and mutated. The expected time
till an $f\!=\!1/f\!=\!2$ individual in $I_1\cup I_2$ is created is
$T_1\approx{1\over \Delta}$ (not ${1\over 2\Delta}$, since only
one coordinate is mutated). From this time on FUSS will select one
half(!) of the time the $f\!=\!1/f\!=\!2$ individual(s) and only the
remaining half the abundant $f\!=\!3$ individuals. When level
$f\!=\!1$ {\em and} level $f\!=\!2$ are occupied, the selection
probability is ${1\over 3}\!+\!{1\over 3}$ for these levels.
With probability $\odt$ the
mutation operator will mutate the $y$ coordinate of an individual
in $I_1$ or the $x$ coordinate of an individual in $I_2$ and
produces a new $f\!=\!1/2/4$ individual. The relative probability
of creating an $f\!=\!4$ individual is $\Delta$. The expected time
to find this global optimum from the $f\!=\!1/f\!=\!2$ individuals, hence,
is $T_2=[({1\over 2}...{2\over 3})\!\times\!{1\over
2}\!\times\!\Delta]^{-1}$. The total expected time is
$T_{FUSS}\approx T_1\!+\!T_2\!=\! {4\over\Delta}...{5\over\Delta}\ll
{1\over\Delta^2}\sim T_{STD}$. FUSS is much faster by exploiting
unfit $f\!=\!1/f\!=\!2$ individuals. This is an example where (local)
minima can help the search. Examples where a low local maxima
can help in finding the global maximum, but where standard
selection sweeps over too quickly to higher but useless local
maxima, can also be constructed.

\paragraph{FUSS with crossover:}
The expected time till an $f\!=\!1$ individual in $I_1$ and an
$f\!=\!2$ individual in $I_2$ is found is $T_1\approx{1\over
\Delta}$, even with crossover. The probability of selecting an
$f=1/f\!=\!2$ individual is ${1\over 3}/{1\over 3}$. Thus, the
probability that a crossing operation crosses $I_1$ with $I_2$ is
$({1\over 3})^2$. The expected time to find the global optimum
from the $f\!=\!1/f\!=\!2$ individuals, hence, is $T_2=9\cdot O(1)$,
where the $O(1)$ factor depends on the frequency of crossover
operations. This is far faster than by STD, even if the
$f\!=\!1/f\!=\!2$ levels were local maxima, since to get a high standard
selection probability, the level has first to be taken over, which
itself needs some time depending on the population size. In FUSS a
single $f\!=\!1$ and a single $f\!=\!2$ individual suffice to
guarantee a high selection probability and an effective crossover.
Crossover does not significantly decrease the {\em total} time
$T_{FUSSX}\approx T_1\!+\!T_2={1\over \Delta}+O(9)$, but for a
suitable 3D generalization we get a large speedup by a factor of
${1\over\Delta}$.

\paragraph{Simple 3D example:}
We generalize the 2D example to D-dimensional individuals
$\vec x\!\in\![0,1]^D$ and a
fitness function
$$
  f(\vec x) \;:=\; (D\!+\!1)\!\cdot\!\prod_{d=1}^D\chi_d(\vec x)\;
  - \max_{1\leq d\leq D} d\!\cdot\!\chi_d(\vec x)\; +d+1,
$$
where $\chi_d(\vec x)$ is the characteristic function of feature
$I_d$
$$
  \chi_d(\vec x) \;:=\; \left\{
  \begin{array}{l}
    1 \quad\mbox{if}\quad a_i\leq x_i\leq a_i+\Delta, \\
    0 \quad\mbox{else.} \\
  \end{array}\right.
$$
For $D\!=\!2$, $f$ coincides with the 2D example. For $D\!=\!3$,
the fractions of $[0,1]^3$ where $f=1/2/3/4/5$ are approximately
$\Delta^2/\Delta^2/\Delta^2/1/\Delta^3$.
With the same line of reasoning we get the following expected
search times for the global optimum:
$$
  T_{RAND}\sim T_{STD}\sim {1\over\Delta^3},
$$ $$
  T_{FUSS}\sim {1\over\Delta^2},\quad
  T_{FUSSX}\sim {1\over\Delta}.
$$
This demonstrates the existence of problems, where FUSS is much
faster than RAND and STD, and that
crossover can give a further boost in FUSS, even when ineffective
in combination with STD.

\paragraph{Numerical results:}
An EA with FUSS and STD has been implemented. First experiments
confirm the superiority of FUSS also for other complicated multimodal
and deceptive fitness functions.
The asymptotic behavior of the convergence times $T$ for
$\Delta\!\to 0$ for the previous example has been verified.
We got similar results for the function $f(x\!+\!a,y\!+\!b)=
2{x^2\over x^2+\Delta^2}+
{y^2\over y^2+\Delta^2}+
4e^{-(x^2+y^2)/\Delta^2}$,
which is a continuous version of the 2D example.
We further applied FUSS to the Traveling Salesman Problem. We
considered $10^{1..3}$ cities with random matrix distances, random
initial paths, random 1-Opt and 2.5-Opt mutation operators, inverse
length as fitness, but no crossover yet. The solutions found by FUSS are
consistently and significantly better than those found by STD. The
current implementation can in no way compete with up-to-date
TSP-solvers \cite{Johnson:97,Applegate:00}, but this was not the
intention of the comparison.
The results are very preliminary yet. Implementation details and
detailed simulation results will be presented in a
forthcoming publication.

\section{Recombination}\label{secCross}

\paragraph{Worst case analysis without recombination:}
We now want to estimate the maximal possible slowdown of FUSS
compared to STD.
Let us assume that all individuals in STD have fitness $f$, and
once one individual with fitness $f\!+\!\eps$ has been found,
takeover of level $f\!+\!\eps$ is quick. Let us assume that this
quick takeover is actually good (e.g.\ if there are no local maxima).
The selection probability of individuals of same fitness is equal.
For FUSS we assume individuals in the range of $f_{min}$ and $f$.
Uniformity is {\em not} necessary. In the worst case, a selection of
an individual of fitness $<\!f$ never leads to an individual of
fitness $\geq\!f$, i.e.\ is always useless. The probability of selecting
an individual with fitness $f$ is $\geq{1\over|F|}$.
At least every $|F|th$ FUSS selection corresponds to a STD
selection. Hence, we expect a maximal slowdown by a factor of
$|F|$, since FUSS ``simulates'' STD statistically every $|F|th$
selection.
It is possible to artificially construct problems where this
slowdown occurs (unimodal function, local mutation
$x\!\to\!x\!\pm\!\eps$, no crossover). We have never observed this
large slowdown in our experiments. For the more complicated
multimodal and deceptive objective functions we were interested
in, FUSS often outperformed STD in solution quality {\em and}
time.

\paragraph{Quadratic slowdown due to recombination:}
We have seen that $T_{FUSS}\!\leq\!|F|\!\cdot\!T_{STD}$. In the
presence of recombination, a {\em pair} of individuals has to be
selected. The probability that FUSS selects {\em two} individuals
with fitness $f$ is $\geq\!{1\over|F|^2}$. Hence, in the worst case,
there could be a slowdown by a factor of $|F|^2$ --- for {\em
independent} selection we expect
$T_{FUSS}\!\leq\!|F|^2\!\cdot\!T_{STD}$. This potential quadratic
slowdown can be avoided by selecting one fitness value at random,
and then two individuals of this single fitness value. For this
{\em dependent} selection, we expect
$T_{FUSS}\!\leq\!|F|\!\cdot\!T_{STD}$. One the other hand,
crossing two individuals of different fitness can also be
advantageous, like the crossing of $f\!=\!1$ with $f\!=\!2$
individuals in the 2D example of Section \ref{secEx}.

\paragraph{Scale independent pair selection:}
It is possible to (nearly) have the best of independent and
dependent selection: a high selection probability $p(f,f')\sim
{1\over|F|}$ if $f\!\approx\!f'$ and $p(f,f')\sim {1\over|F|^2}$
otherwise, with uniform marginal $p(f)\!=\!{1\over|F|}$. The idea
is to use a strongly correlated joint distribution for selecting a
fitness pair. A ``scale independent'' probability distribution
$p(f,f')\sim{1\over|f-f'|}$ is appropriate. We define the joint
probability $\tilde p(f,f')$ of selecting two individuals of
fitness $f$ and $f'$ and the marginal $\tilde p(f)$ as
\beq\label{ptjoint}
  \tilde p(f,f') \;:=\; {1\over 2|F|\ln|F|}\cdot
  {1\over {1\over\eps}|f\!-\!f'|+1},
\eeq
$$
  \tilde p(f) \;:=\; \sum_{f'\in F}\tilde p(f,f')
  = \sum_{f'\in F}\tilde p(f',f).
$$
We assume $|F|\geq 3$ in the following. The $+1$ in the
denominator has been added to regularize the expression for
$f\!=\!f'$. The factor $(2|F|\ln|F|)^{-1}$ ensures correct
normalization for $|F|\!\to\!\infty$. More precisely, using
$\ln{b+1\over a}\leq\sum_{i=a}^b{1\over i}\leq\ln{b\over a-1}$,
one can show that
$$
  1-{\textstyle{1\over\ln|F|}} \;\leq
  \sum_{f,f'\in F}\tilde p(f,f') \;\leq\; 1
  ,\quad
  \odt \;\leq\; |F|\!\cdot\!\tilde p(f) \;\leq\; 1
$$
i.e.\ $\tilde p$ is not strictly normalized to $1$ and the
marginal $\tilde p(f)$ is only approximately (within a factor of 2)
uniform. The first defect can be corrected by appropriately
increasing the diagonal probabilities $\tilde p(f,f)$. This also
solves the second problem.
\beq\label{pjoint}
  p(f,f') \;:=\; \left\{
  \begin{array}{ll}
    \tilde p(f,f') & \mbox{for}\quad f\neq f' \\
    \tilde p(f,f')+[{1\over|F|}-\tilde p(f)] &
    \mbox{for}\quad f=f' \
  \end{array}
\right.
\eeq

\paragraph{Properties of $p(f,f')$:}
$p$ is normalized to $1$ with uniform marginal
$$
  p(f):= \sum_{f'\in F} p(f,f') = {1\over|F|},\quad
  p(f,f')\geq \tilde p(f,f'),
$$ $$
  \sum_{f,f'\in F} p(f,f') =
  \sum_{f\in F} p(f) = 1,
$$
Apart from a minor additional logarithmic suppression of order
$\ln|F|$ we have the desired behaviour $p(f,f')\sim {1\over|F|}$
for $f\approx f'$ and $p(f,f')\sim {1\over|F|^2}$ otherwise:
$$
  p(f,f\pm m\eps) \geq {1\over 2\ln|F|} \cdot
  {1\over m+1} \cdot {1\over|F|},
$$ $$
  p(f,f') \geq {1\over 2\ln|F|} \cdot
  {1\over |F|^2}.
$$
During optimization, the minimal/maximal fitness of an
individual in population $P_t$ is $f_{min/max}^t$. In the
definition of $p$ one has to use
$F_t\!:=\!\{f_{min}^t,f_{min}^t\!+\!\eps,...,f_{max}^t\}$ instead
of $F$, i.e.\ $|F|\hookrightarrow
|F_t|={1\over\eps}(f_{max}^t\!-\!f_{min}^t)+1\leq|F|$.

\section{Continuous Fitness Functions}\label{secCont}
\paragraph{Effective discretization scale:}
Up to now we have considered a discrete valued fitness function
with values in $F\!=\!\{f_{min},f_{min}\!+\!\eps,...,f_{max}\}$.
In many practical problems, the fitness function is continuous
valued with $F=[f_{min},f_{max}]$. We generalize FUSS, and some of
the discussion of the previous sections to the continuous case by
replacing the discretization scale $\eps$ by an effective
(time-dependent) discretization scale $\hat\eps$. By construction,
FUSS shifts the population towards a more uniform one. Although
the fitness values are no longer equi-spaced, they still form a
discrete set for finite population $P$. For a fitness uniform
distribution, the average distance between (fitness) neighboring
individuals is
${1\over|P_t|-1}(f^t_{max}\!-\!f^t_{min})\!=:\!\hat\eps$. We
define $\hat
F_t\!:=\!\{f^t_{min},f^t_{min}\!+\!\hat\eps,...,f^t_{max}\}$.
$|\hat F_t| = {1\over\hat\eps}(f^t_{max}\!-\!f^t_{min})+1 =
|P_t|$.

\paragraph{FUSS:}
Fitness uniform selection for a continuous valued function has
already been mentioned in Section \ref{secFuss}. We just take a
uniform random fitness $f$ in the interval
$[f_{min}^t-\odt\hat\eps,f_{max}^t+\odt\hat\eps]$. One may even
take the limit $\hat\eps\!\to\!0$ in this case probably without
harming FUSS.
Independent and dependent fitness pair selection as
described in the last section works analogously. An
$\hat\eps\!=\!0$ version of correlated selection does not exist; a
non-zero $\hat\eps$ is important. A discrete pair $(f,f')$ is
drawn with probability $p(f,f')$ as defined in
(\ref{ptjoint}) and (\ref{pjoint}) with $\eps$ and $F$ replaced by
$\hat\eps$ and $\hat F_t$. The additional suppression $\ln|\hat
F_t|\!=\!\ln|P_t|$ is small for all practical population sizes.

In all cases an individual with fitness nearest to $f$ ($f'$) is
selected from the population $P$ (randomly if there is
more than one nearest individual).

\paragraph{Discussion:}
If we assume a fitness uniform distribution, a worst case bound
$T_{FUSS}\approxleq\sum_{t=1}^{T_{STD}}|P_t|$ seems plausible,
since the probability of selecting the best individual is
approximately $|P_t|$. For constant population size we get a bound
$T_{FUSS}\approxleq|P|\!\cdot\!T_{STD}$. For the preferred
non-deletion case with population size $|P_t|\!=\!t$ the bound
gets much worse $T_{FUSS}\approxleq\odt T_{STD}^2$.
This possible (but not necessary!) slowdown has similarities to
the slowdown problems of proportionate selection in later
optimization stages.
Larger choices of $\hat\eps$ may be favorable if the standard
choice causes problems.

\section{Summary \& Conclusions}\label{secConc}
We have addressed the problem of balancing the selection intensity
in EAs, which determines speed versus quality of a solution. We
invented a new fitness uniform selection scheme FUSS. It generates
a selection pressure towards sparsely populated fitness levels.
This property is unique to FUSS as compared to other selection
schemes (STD).
It results in the desired high selection pressure towards higher
fitness if there are only a few fit individuals. The selection
pressure is automatically reduced when the number of fit
individuals increases.
A joint pair selection scheme for recombination has been
defined, but not yet implemented.
A heuristic worst case analysis of FUSS compared to STD has been
given.
FUSS solves the problem of exponential takeover and the resulting
loss of genetic diversity of STD, while still generating enough
selection pressure. It does not help in getting a more uniform
distribution within a fitness level. We showed analytically by way
of a simple example that FUSS can be much more effective than STD.
FUSS should be compared to STD on other problems to further
explore its efficacy and limitations. First results look
encouraging.
Of special interest is whether FUSS could improve up-to-date
EAs that solve difficult combinatoric optimization problems, like
TSPs. We expect FUSS to be superior to STD in cases where an EA
with STD effectively gets trapped into local optima.


\end{document}